\documentclass[twoside]{article}

\usepackage[accepted]{aistats2020}

\usepackage[utf8]{inputenc} 
\usepackage[T1]{fontenc}    
\usepackage{hyperref}       
\usepackage{url}            
\usepackage{booktabs}       
\usepackage{amsfonts}       
\usepackage{nicefrac}       
\usepackage{microtype}      

\usepackage{graphicx}
\usepackage{latexsym}
\usepackage{url}
\usepackage{amssymb}
\usepackage{amsmath}
\usepackage{amsthm}
\usepackage{booktabs}
\usepackage{epstopdf}
\usepackage{subfigure} 
\usepackage{color}
\usepackage{todonotes}

\begin{document}

%

%

\twocolumn[

\aistatstitle{The Indian Chefs Process}

\aistatsauthor{Patrick Dallaire\textsuperscript{1}  \And Luca Ambrogioni\textsuperscript{2} \And  Ludovic Trottier\textsuperscript{1}
\And Umut Güçlü\textsuperscript{2} \And Max Hinne\textsuperscript{2}
}
\aistatsauthor{
Philippe Giguère\textsuperscript{1} \And Brahim Chaib-Draa\textsuperscript{1}
\And Marcel van Gerven\textsuperscript{2}  \And Francois Laviolette\textsuperscript{1}
}
\aistatsaddress{ \textsuperscript{1}Université Laval \And \textsuperscript{2}Radboud University } ]

\begin{abstract}
This paper introduces the Indian Chefs Process (ICP), a Bayesian nonparametric prior on the joint space of infinite directed acyclic graphs (DAGs) and orders that generalizes Indian Buffet Processes. As our construction shows, the proposed distribution relies on a latent Beta Process controlling both the orders and outgoing connection probabilities of the nodes, and yields a probability distribution on sparse infinite graphs. The main advantage of the ICP over previously proposed Bayesian nonparametric priors for DAG structures is its greater flexibility. To the best of our knowledge, the ICP is the first Bayesian nonparametric model supporting every possible DAG. We demonstrate the usefulness of the ICP on learning the structure of deep generative sigmoid networks as well as convolutional neural networks.   
\end{abstract}

\section{Introduction}

In machine learning and statistics, the directed acyclic graph~(DAG) is a common modelling choice for expressing relationships between objects. Prime examples of DAG-based graphical models include Bayesian networks, feed-forward neural networks, causal networks, deep belief networks, dynamic Bayesian networks and hidden Markov models, to name a few. Learning the unknown structure of these models presents a significant learning challenge, a task that is often avoided by fixing the structure to a large and hopefully sufficiently expressive model. 
\textit{Structure learning} is a model selection problem in which ones estimates the underlying graphical structure of the model. Over the years, researchers have explored a great variety of approaches to this problem~\cite{jordan1998learning,schmidt2007learning,banerjee2015bayesian, kwok1997constructive, mansinghka2012structured, mohammadi2015bayesian, tervo2016toward}, from frequentist to Bayesian, and some using pure heuristic-based search, but the vast majority is limited to finite parametric models. 

Bayesian nonparametric learning methods are appealing alternatives to their parametric counterparts, because they offer more flexibility when dealing with generative models of unknown dimensionality~\cite{hjort2009invitation}. Instead of looking for specific finite-dimensional models, the idea is rather to define probability measures on infinite-dimensional spaces and then infer the finite subset of active dimensions explaining the data. Over the past years, there has been extensive work on constructing flexible Bayesian nonparametric models for various types of graphical models, allowing complex hidden structures to be learned from data. For instance,~\cite{jiang2013infinite} developed a model for infinite latent conditional random fields while~\cite{nbn2010} proposed an infinite mixture of fully observable finite-dimensional Bayesian networks. In the case of time series,~\cite{chatzis2010infinite} developed the infinite hidden Markov random field model and~\cite{doshi2011infinite} proposed an infinite dynamic Bayesian network with factored hidden states. Another interesting model is the infinite factorial dynamical model of~\cite{valera2015infinite} representing the hidden dynamics of a system with infinitely many independent hidden Markov models.

The problem of learning networks containing hidden structures with Bayesian nonparametric methods has also received attention. The cascading Indian Buffet Process (CIBP) of~\cite{adams2010a} is a Bayesian nonparametric prior over infinitely deep and infinitely broad layered network structures. However, the CIBP does not allow connections from non-adjacent layers, yielding a restricted prior over infinite DAGs. The extended CIBP (ECIBP) is an extension of the previous model which seeks to correct this limitation and support a larger set of DAG structures~\cite{dallaire2014learning}. However, the ECIBP has some drawbacks: the observable nodes are confined to a unique layer placed at the bottom of the network, which prevents learning the order of the nodes or have observable inputs. An immediate consequence of this is the impossibility for an observable unit to be the parent of any hidden unit or any other observable unit, which considerably restricts the support of the prior over DAGs and make their application to deep learning very problematic. 

In the context of deep learning, structure learning is often part of the optimization. Recently, \cite{wen2016learning} proposed a method that enforces the model to dynamically learn more compact structures by imposing sparsity through regularization. While sparsity is obviously an interesting property for large DAG-based models, their method ignores the epistemic uncertainty about the structure. Structure learning for probabilistic graphical models can also be applied in deep learning. For instance, \cite{rohekar2018constructing} have demonstrated that deep network structures can be learned throught the use of Bayesian network structure learning strategies. To our knowledge, no Bayesian nonparametric structure learning methods have been applied to deep learning models. 

This paper introduces the Indian Chefs Process (ICP), a new Bayesian nonparametric prior for general DAG-based structure learning, which can equally be applied to perform Bayesian inference in probabilistic graphical models and deep learning. The proposed distribution has a support containing all possible DAGs, admits hidden and observable units, is layerless and enforces sparsity. We present its construction in Section \ref{sec:probability_dag} and describe a learning method based on Markov Chain Monte Carlo in Section \ref{sec:structure_learning}. In Section \ref{sec:learning_experiments}, we use the ICP as a prior in two Bayesian structure learning experiments: in the first, we compute the posterior distribution on the structure and parameters of a deep generative sigmoid networks and in the second we perform structure learning in convolutional neural networks.

\section{Bayesian nonparametric directed acylic graphs}
\label{sec:probability_dag}

In this section, we construct the probability distribution over DAGs and orders by  adopting the methodology followed by~\cite{griffiths2006infinite}. We first define a distribution over finite-dimensional structures, while the final distribution is obtained by evaluating it as the structure size grows to infinity. 

Let $G = (V,Z)$ be a DAG where $V = \{1,\dots,K\}$ is the set of nodes and $Z \in \{0,1\}^{K \times K}$ is the adjacency matrix. We introduce an ordering $\boldsymbol\theta$ on the nodes so that the direction of an edge is determined by comparing the order value of each node. A connection $Z_{ki}=1$ is only allowed when $\theta_k > \theta_i$, meaning that higher order nodes are parents and lower order nodes are children. Notice that this constraint is stronger than acyclicity since all $(Z,\boldsymbol\theta)$ combinations respecting the order value constraint are guaranteed to be acyclic, but an acyclic graph can violate the ordering constraint. 

We assume that both the adjacency matrix $Z$ and the ordering $\boldsymbol\theta$ are random variables and develop a Bayesian framework reflecting our uncertainty. Accordingly, we assign a popularity parameter $\pi_k$ and an order value $\theta_k$, called reputation, to every node $k$ in $G$ based on the following model:
\begin{align}
    \theta_k &\thicksim \mathcal{U}(0,1) \label{eq:uniform_order}\\
    \pi_k \mid \alpha, \gamma, \phi, K &\thicksim \text{Beta}\left(\frac{\alpha\gamma}{K} + \phi  \mathbb{I}(k \in O), \alpha - \frac{\alpha\gamma}{K}\right)  \label{eq:beta_prior}\\
    Z_{ki} \mid \pi_k, \theta_k, \theta_i &\thicksim \text{Bernoulli} \left(\pi_k  \mathbb{I}(\theta_k > \theta_i) \right)~.
\end{align} 
Here, $\mathbb{I}$ denotes the indicator function, $\mathcal{U}(a,b)$ denotes the uniform distribution on interval $[a,b]$ and $O \subseteq V$ is the set of \emph{observed} nodes. In this model, the popularities reflected by $\pi$ control the outgoing connection probability of the nodes, while respecting the \emph{total order} imposed by $\boldsymbol\theta$. Moreover, the Beta prior parametrization in Eq.~\eqref{eq:beta_prior} is motivated by the Beta Process construction of~\cite{paisley10b}, where Eq.~\eqref{eq:uniform_order} becomes the \textit{base distribution}, and is convenient when evaluating the limit in Section~\ref{sec:finite_to_infinite}. Also, $\alpha$ and $\gamma$ correspond to the usual parameters defining a Beta Process and the purpose of the new parameter $\phi$ is to control the popularity of the observable nodes and ensure a non-zero connection probability when required.

Under this model, the conditional probability of the adjacency matrix $Z$ given the popularities  $\boldsymbol\pi = \{\pi_k\}_{k=1}^K$ and order values $\boldsymbol\theta = \{\theta_k\}_{k=1}^K$ is:
\begin{align}\label{eq:p_Z_given_pi}
    p(Z \mid \boldsymbol\pi, \boldsymbol\theta) &= \prod_{k=1}^K \prod_{i=1}^{K} p(Z_{ki}\mid \pi_k, \theta_k, \theta_i) ~.
\end{align}
The adjacency matrix $Z$ may contain connections for nodes that are not of interest, i.e. nodes that are not ancestors of any observable nodes. Formally, we define $A \subseteq V$ as the set of \emph{active} nodes, which contains all observable nodes $O$ and the ones having a directed path ending at an observable node. 

When solely considering connections from $A$ to $A$, i.e. the adjacency submatrix $Z_{AA}$ of the $A$-induced subgraph of $G$, Eq.~\eqref{eq:p_Z_given_pi} simplifies to:
\begin{align}\label{eq:Z_AA}
    p(Z_{AA} \mid \boldsymbol\pi, \boldsymbol{\downarrow}, \boldsymbol\theta) = \prod_{k \in A} \pi_k^{m_k} \left(1- \pi_k \right)^{\downarrow_k - m_k}~,
\end{align}
where $m_k= \sum_{i \in A} Z_{ki}$ denotes the number of outgoing connections from node $k$ to any active nodes, \mbox{$\downarrow_k = \sum_{j \in A} \mathbb{I}(\theta_j < \theta_k)$} denotes the number of active nodes having an order value strictly lower than $\theta_k$ and $\boldsymbol\downarrow\,= \{\downarrow_k\}_{k=1}^K$. At this point, we marginalize out the popularity vector $\boldsymbol\pi$ in Eq.~\eqref{eq:Z_AA} with respect to the prior, by using the conjugacy of the Beta and Binomial distributions, and we get:
\begin{align}\label{eq:Z_AA_marginal}
p(Z_{AA} \mid  \alpha, \gamma, &\phi, \boldsymbol{\downarrow}, \boldsymbol\theta) = \nonumber \\
&\prod_{k \in H} \frac{[ \frac{\alpha\gamma}{K}]^{\overline{m_k}} [\alpha - \frac{\alpha\gamma}{K}]^{\overline{\downarrow_k - m_k}}}{\alpha^{ \overline{\downarrow_k} }} \\
& \prod_{k \in O} \frac{[ \frac{\alpha\gamma}{K} + \phi]^{\overline{m_k}} [\alpha - \frac{\alpha\gamma}{K}]^{\overline{\downarrow_k - m_k}}}{[\alpha + \phi]^{ \overline{\downarrow_k} } }, \nonumber 
\end{align}
where $x^{\overline{n}} = x(x+1) \dots (x + n -1)$ is the Pochhammer symbol denoting the rising factorial and $H = A \setminus O$ is the set of active hidden nodes.

The set of active nodes $A$ contains all observable nodes as well as their ancestors, which means there exists a part of the graph $G$ that is disconnected from $A$. Let us denote by $I = V \setminus A$ the set of \emph{inactive} nodes. Considering that the $A$-induced subgraph is effectively maximal, then this subgraph must be properly isolated by some envelope of no-connections $Z_{IA}$ containing only zeros. The joint probability of submatrices $Z_{AA}$ and $Z_{IA}$ is: 
\begin{align}\label{eq:joint_proba_dags}
p(Z_{AA}, &Z_{IA} \mid \alpha, \gamma, \phi, \boldsymbol{\downarrow}, \boldsymbol\theta) = \nonumber \\
& p(Z_{AA} \mid \alpha, \gamma, \phi, \boldsymbol{\downarrow}, \boldsymbol\theta) \cdot \prod_{k \in I} \frac{ [\alpha - \frac{\alpha\gamma}{K}]^{\overline{\downarrow_k}}}{\alpha^{ \overline{\downarrow_k} }}
\end{align}
where the number of negative Bernoulli trials $\downarrow_k$ depends on $\theta_k$ itself and $\boldsymbol\theta_A$. Notice that since the submatrices $Z_{AI}$ and $Z_{II}$ contain uninteresting and unobserved binary events, they are trivially marginalized out of $p(Z)$.

One way to simplify Eq.~\eqref{eq:joint_proba_dags} is to marginalize out the order values $\boldsymbol\theta_I$ of the inactive nodes with respect to \eqref{eq:uniform_order}. To do so, we first sort the active node orders ascendingly in vector $\boldsymbol\theta_A^\nearrow$ and augment it with the extrema $\theta_{0}^\nearrow = 0$ and $\theta_{K^+ + 1}^\nearrow = 1$, where we introduce $K^+ = |A|$ to denote the number of active nodes. We slightly abuse notation here since these extrema do not refer to any nodes and are only used to compute interval lengths. This provides us with all relevant interval boundaries, including the absolute boundaries implied by Eq.~\eqref{eq:uniform_order}. We refer to the $j$\textsuperscript{th} smallest value of this vector as $\theta_j^\nearrow$. Based on the previous notation, the probability for an inactive node to lie between two active nodes is simply $\theta_{j+1}^\nearrow - \theta_j^\nearrow$. Using this notation, we have the following marginal probability:
\begin{align}
p(&Z^\nearrow_{AA}, Z_{IA}, \boldsymbol\theta_A^\nearrow  \mid \alpha, \gamma, \phi) = \nonumber \\
 &\frac{(K-D)^{\underline{ K^+ - D}}}{K^+!} 
  \left( \sum_{j=0}^{K^+} (\theta_{j+1}^\nearrow - \theta_{j}^\nearrow) \frac{[\alpha(1 - \frac{\gamma}{K})]^{\overline{j}}}{\alpha^{ \overline{j} }} \right)^{K^-}   \nonumber \\
&\prod_{k \in H} \frac{[ \frac{\alpha\gamma}{K}]^{\overline{m_k}} [\alpha - \frac{\alpha\gamma}{K}]^{\overline{\downarrow_k - m_k}}}{\alpha^{ \overline{\downarrow_k} }} \label{eq:structure_marginal_joint_proba_dags} \\ 
&\prod_{k \in O} \frac{[ \frac{\alpha\gamma}{K} + \phi]^{\overline{m_k}} [\alpha - \frac{\alpha\gamma}{K}]^{\overline{\downarrow_k - m_k}}}{[\alpha + \phi]^{ \overline{\downarrow_k} } },   \nonumber
\end{align}
where $K^- = |I|$ denotes the number of inactive nodes and $x^{\underline{n}} = x(x-1) \dots (x - n + 1)$ symbolizes the falling factorial. Due to the exchangeability of our model, the joint probability on both the adjacency matrix and active order values can cause problems regarding the index $k$ of the nodes. One way to simplify this is to reorder the adjacency matrix according to $\boldsymbol\theta_A^\nearrow$, which we denote $Z^\nearrow_{AA}$. By using this many-to-one transformation, we obtain a probability distribution on an equivalence class of DAGs that is analog to the \emph{lof} function used by~\cite{griffiths2006infinite}. The number of permutation mapping to this sorted representation is accounted for by the normalization constant $(K-D)^{\underline{ K^+ - D}}(K^+!)^{-1}$. 

\subsection{From Finite to Infinite DAGs}\label{sec:finite_to_infinite}

An elegant way to construct Bayesian nonparametric models is to consider the infinite limit of finite parametric Bayesian models~\cite{orbanz2010bayesian}. Following this idea, we revisit the model of Section~\ref{sec:probability_dag} so that $G$ now contains infinitely many nodes. To this end, we evaluate the limit as $K \rightarrow \infty$ of Eq.~\eqref{eq:structure_marginal_joint_proba_dags}, yielding the following probability distribution:
\begin{align}
p(&Z_{AA}^\nearrow, Z_{IA} , \boldsymbol\theta_A^\nearrow | \alpha, \gamma, \phi, O) = \nonumber \\
&\frac{1}{K^+!} \exp \left( -\alpha\gamma \sum_{j=1}^{K^+}(\theta_{j+1}^\nearrow - \theta_j^\nearrow) \big{[} \psi(\alpha+j) - \psi(\alpha) \big{]} \right) \nonumber  \\
&\prod_{k \in H}  \alpha\gamma \frac{ (m_k-1)!}{(\alpha + \downarrow_k - m_k)^{\overline{m_k}}}
\prod_{k \in O} \frac{ \phi^{\overline{m_k}} \alpha^{\overline{\downarrow_k - m_k}}}{[\alpha + \phi]^{ \overline{\downarrow_k} } }, 
\label{eq:marginal_joint_proba_infinite_dags}
\end{align}
where $\psi$ is the digamma function. Eq.~\eqref{eq:marginal_joint_proba_infinite_dags} is the proposed marginal probability distribution on the joint space of infinite DAGs and continuous orders. The Indian Chefs Process\footnote{Described in the supplementary material along with some properties of the distribution. This content is not necessary to reproduce the experimental results.} is a stochastic process for generating random DAGs from \eqref{eq:marginal_joint_proba_infinite_dags}, making it an equivalent representative of this distribution.  

\subsection{The Indian Chefs Process}\label{sec:sampling}

Now that we have the marginal probability distribution~\eqref{eq:marginal_joint_proba_infinite_dags}, we want to draw random active subgraphs from it. This section introduces the Indian chefs process (ICP), a stochastic process serving this purpose. In the ICP metaphor, chefs draw inspiration from other chefs, based on their \emph{popularity} and \emph{reputation}, to create the menu of their respective restaurant. This creates inspiration maps representable with directed acyclic graphs. ICP defines two types of chefs: 1) star chefs (corresponding to observable variables) which are introduced iteratively and 2) regular chefs (corresponding to hidden variables) which appear only when another chef selects them as a source of inspiration. 

The ICP starts with an empty inspiration map as its initial state. The infinitely many chefs can be thought of as lying on a unit interval of reputations. Every chef has a fraction of the infinitely many chefs above him and this fraction is determined by the chef's own reputation.

The general procedure at iteration $t$ is to introduce a new star chef, denoted $i$, within a fully specified map of inspiration representing the connections of the previously processed chefs. The very first step is to draw a reputation value from $\theta_i \thicksim \mathcal{U}(0, 1)$ to determine the position of the star chef in the reputation interval. Once chef $i$ is added, sampling the new inspiration connections is done in three steps. 

\paragraph{Backward proposal}
Step one consists in proposing \textit{star} chef $i$ as an inspiration to \textit{all} the $\downarrow_i$ chefs having a lower reputation than chef $i$. To this end, we can first sample the total number of inspiration connections with: 
\begin{equation}
q_i \thicksim \text{Binomial}\left(\downarrow_i , \frac{\phi}{\alpha + \phi}\right),
\end{equation}
and then uniformly pick one of the $\binom{\downarrow_i}{q_i}$ possible configurations of inspiration connections. 

\paragraph{Selecting existing chefs}
In step two, chef $i$ considers \textit{any} already introduced chefs of higher reputation. The probability for candidate chef $k$ to become an inspiration for $i$ is:
\begin{equation}
Z_{ki} \thicksim \text{Bernoulli}\left(\frac{m_k + \phi\mathbb{I}(k \in \text{star chefs})}{\alpha + \downarrow_k - 1 + \phi \mathbb{I}(k \in \text{star chefs})}\right),
\end{equation}
where $\downarrow_k$ includes the currently processed chef $i$.

\paragraph{Selecting new chefs}
The third step allows chef $i$ to consider completely new \textit{regular} chefs as inspirations in every single interval above $i$. The number of new regular chefs $K_j^{new}$ to add in the $j$\textsuperscript{th} reputation interval above $i$ follows probability distribution:
\begin{equation}
K^{new}_{j} \thicksim \text{Poisson}\left(\frac{ (\theta^\nearrow_{j+1} - \theta^\nearrow_{j})  \alpha \gamma  }{\alpha + \downarrow_j - 1}   \right),
\end{equation}
where the new regular chefs are independently assigned a random reputation drawn from $\mathcal{U}(\theta^\nearrow_{j}, \theta^\nearrow_{j+1})$. The \textit{regular} chefs introduced during this step will be processed one by one using step two and three. Once all newly introduced regular chefs have been processed, the next iteration $t+1$ can begin with step one, a step reserved to star chefs only.

\subsection{Some properties of the distribution}
\label{sec:properties}

\begin{figure*}[ht!]
\centering
\subfigure[]{\label{fig:alpha_gamma}
    \includegraphics[trim = 8mm 0mm 13.5mm 5mm, clip,scale=0.25]{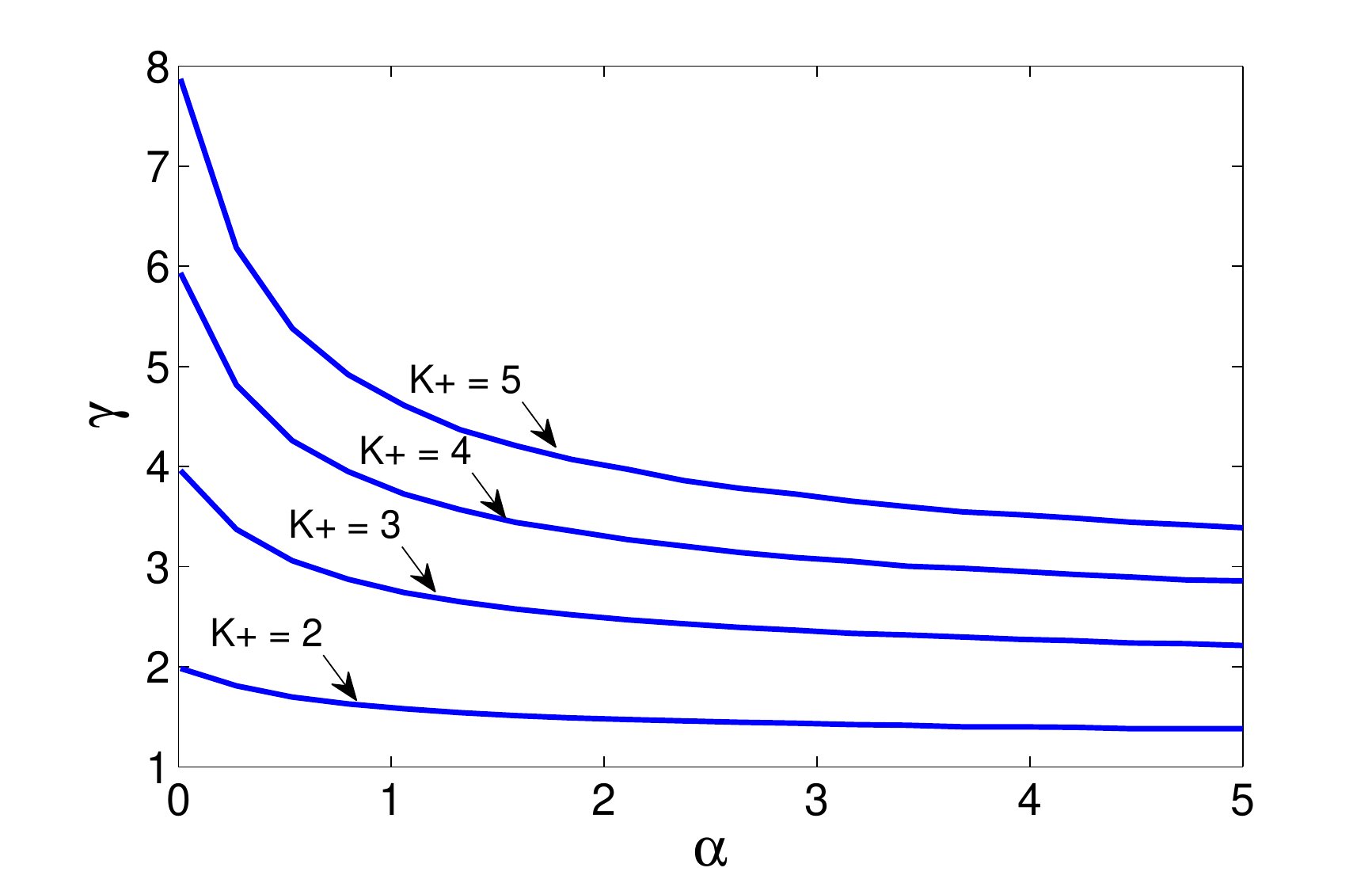}
}
\subfigure[]{\label{fig:alpha_E_plus}
    \includegraphics[trim = 8mm 0mm 13.5mm 5mm, clip,scale=0.25]{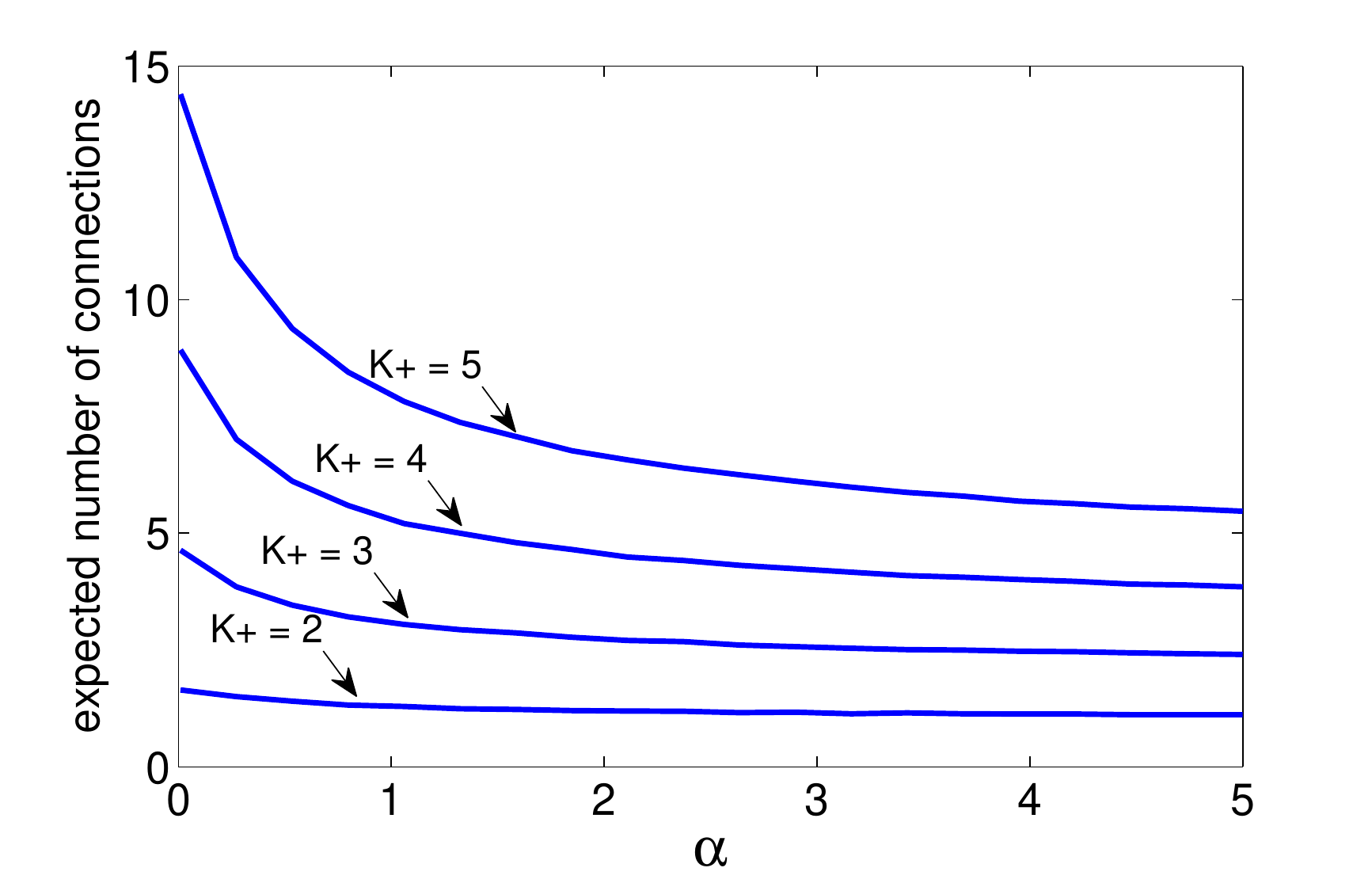}
}
\subfigure[]{\label{fig:N_alpha}
    \includegraphics[trim = 8mm 0mm 13.5mm 5mm, clip,scale=0.25]{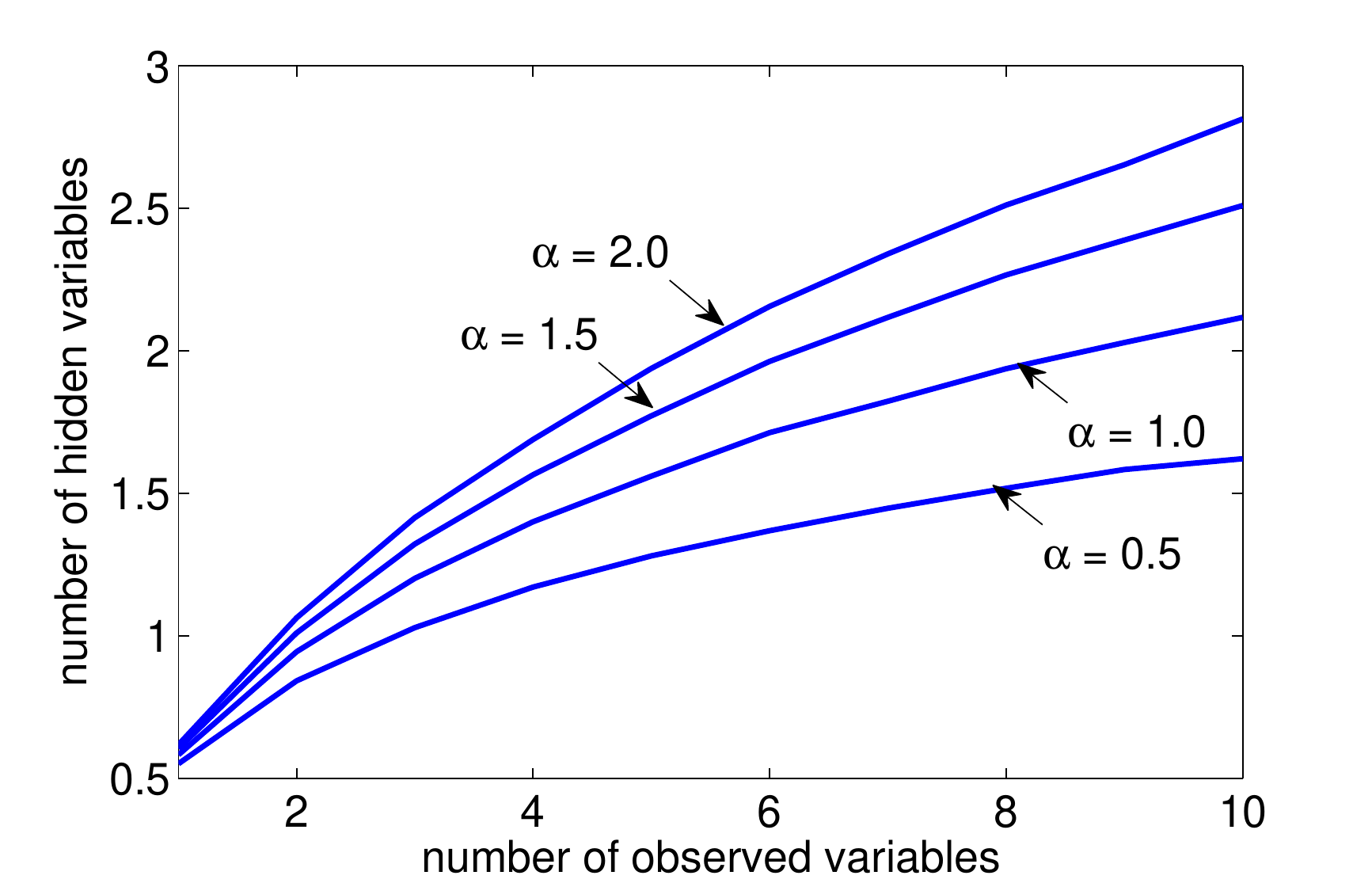}
}
\caption{Empirical study of hyperpameters. Figure (a) shows the expected number of active nodes as a function of $\alpha$ and $\gamma$. Figure (b) shows that once we know the expected $K^+$ from $\alpha$ and $\gamma$, we can find the expected number of connections. Figure (c) shows the influence of $\alpha$ (with $\gamma = 1$) on the complexity (number of hidden nodes) as function of the number of observable nodes.}
\label{fig:dist_properties_alpha_gamma}
\end{figure*}

To better understand the effect of the hyperparameters on the graph properties, we performed an empirical study of some relations between the hyperparameters, the expected number of active nodes $\mathbb{E}[K^+ | \alpha, \gamma]$ and the expected number of active edges $\mathbb{E}[E^+ | \alpha, \gamma]$, where $E^+$ is the number of elements in $Z_{AA}$. Figure~\ref{fig:alpha_gamma} depicts level curves of $\mathbb{E}[K^+ | \alpha, \gamma]$ for the case of only 1 observable placed at $\theta_k = 0$. The figure shows that several combinations of $\alpha$ and $\gamma$ leads to the same expected number of active nodes. Notice that fixing one hyperparameter, either $\alpha$ or $\gamma$, and selecting the expected number of nodes, one can retrieve the second hyperparameter that matches the relationship. We used this fact in the construction of Figure \ref{fig:alpha_E_plus} where the unshown parameter $\gamma$ could be calculated. In Figure \ref{fig:alpha_E_plus}, we illustrate the effect of $\alpha$ on $\mathbb{E}[E^+ | \alpha, \gamma]$ which essentially shows that smaller values of $\alpha$ increase the graph density. 

When using Bayesian nonparametric models, we are actually assuming that the generative model of the data is infinite-dimensional and that only a finite subset of the parameters are involved in producing a finite set of data. The effective number of parameters explaining the data corresponds to the model complexity and usually scales logarithmically with respect to the sample size. Unlike most Bayesian nonparametric models, the ICP prior scales according to the number of observed nodes added to the network. In Figure~\ref{fig:dist_properties_alpha_gamma}, we show how the expected number of active hidden nodes increases as function of the number of observable nodes. 

\subsection{Connection to the Indian Buffet Process}

There exists a close connection between the Indian Chefs Process (ICP) and the Indian Buffet Process (IBP). In fact, our model can be seen as a generalization of the IBP. Firstly, all realizations of the IBP receive a positive probability under the ICP. Secondly, the two-parameter IBP~\cite{griffiths2011indian} is recovered, at least conceptually, when altering the prior on order values (see Eq.~\eqref{eq:uniform_order}) so that all observed nodes are set to reputation $\theta = 0$ and all hidden nodes are set to reputation $\theta = 1$. This way, connections are prohibited between hidden nodes and between observable nodes, while hidden-to-observable connections are still permitted. 

\section{Structure Learning}
\label{sec:structure_learning}

In this section, we present a Markov Chain Monte Carlo (MCMC) algorithm approximating the exact ICP prior distribution with samples. We propose a reversible jump MCMC algorithm producing random walks on Eq.~\eqref{eq:marginal_joint_proba_infinite_dags}~\cite{green2009reversible}. This algorithm works in three phases: the first resamples graph connections without adding or removing any nodes, the second phase is a birth-death process on nodes and the third one only involves the order. 

The algorithm itself uses the notion of \emph{singleton} and \emph{orphan} nodes. A node is a singleton when it only has a unique active child. Thus, removing its unique connection would disconnect the node from the active subgraph. Moreover, a node is said to be an orphan if it does not have any parents. 

\paragraph{Within model moves on adjacency matrix:} We begin by uniformly selecting a node $i$ from the active subgraph. Here, the set of parents to consider for $i$ comprises all non-singleton active nodes having an order value greater than $\theta_i$. This set includes both current parents and candidate parents. Then, for each parent $k$, we Gibbs sample the connections using the following conditional probability:
\begin{equation}
p(Z^\nearrow_{ki} = 1 | Z^{^\nearrow\neg ki}_{AA}, \boldsymbol\theta_A)  = \frac{m_{k}^{\neg i} + \phi  \mathbb{I}(k \in O)}{\alpha + \downarrow_k - 1  + \phi \mathbb{I}(k \in O)} 
\, ,
\end{equation}
where $m_{k}^{\neg i}$ is the number of outgoing connections of node $k$ excluding connections going to node $i$ and $Z^{^\nearrow\neg ki}_{AA}$ has element $ki$ removed. Also, all connections not respecting the order are prohibited and therefore have an occurrence probability of 0, along with (trans-dimentional) singleton parent moves.

\paragraph{Trans-dimensional moves on adjacency matrix:} We begin with a random uniform selection of node $i$ in the active subgraph and, with equal probability, propose either a \textit{birth} or a \textit{death} move. 

In the birth case, we activate node $k$ by connecting it to node $i$. Its order $\theta_k$ is determined by uniformly selecting an insertion interval above $\theta_i$. Assuming node $i$ is also the $i$\textsuperscript{th} element in $\boldsymbol\theta^\nearrow_{A}$, we have $\uparrow_i = K^+ - i + 1$ possible intervals, including zero-length intervals. Let us assume that $j$ and $j+1$ are the two nodes between which $k$ is to be inserted. Then, we obtain the candidate order value of the new node by sampling $\theta_k \thicksim \mathcal{U}(\theta_{j}^\nearrow, \theta_{j+1}^\nearrow)$. The Metropolis-Hastings acceptance ratio here is:
\begin{equation}
\begin{split}
a_{birth} = \min  \Bigg{\{} 1, &\frac{p({Z'}_{A'A'}^\nearrow, Z'_{I'A'} , {\boldsymbol\theta'}_{A'}^\nearrow | \alpha, \gamma, \phi, O )}{p(Z^\nearrow_{AA}, Z_{IA} , \boldsymbol\theta^\nearrow_A | \alpha, \gamma, \phi, O )} \\ 
&\cdot \frac{ (\theta_{j+1}^\nearrow - \theta_{j}^\nearrow)(\uparrow_i + 1)K^+}{K^*_i + 1}
\Bigg{\}},
\end{split}
\end{equation}
where $K^*_i$ is the number of singleton-orphan parents of $i$ and $\uparrow_i = \sum_{j \in A} \mathbb{I}(\theta_j > \theta_i)$ is the number of active nodes above~$i$.

In the death case, we uniformly select one of the $K^*_i$ singleton-orphan parents of $i$  if $K^*_i > 0$ and simply do nothing in case there exists no such node. Let $k$ be the parent to disconnect and consequently deactivate. The Metropolis-Hastings acceptance ratio for this move is:
\begin{equation}
\begin{split}
a_{death} = \min \Bigg{\{} 1,  &\frac{p({Z'}_{A'A'}^\nearrow, Z'_{I'A'} , {\boldsymbol\theta'}_{A'}^\nearrow | \alpha, \gamma, \phi, O )}{p(Z_{AA}^\nearrow, Z_{IA} , \boldsymbol\theta^\nearrow_A | \alpha, \gamma, \phi, O )} \\ &\cdot \frac{K^*_i } {(\theta_{j+1}^\nearrow - \theta_{j}^\nearrow) (K^+ - 1) \uparrow_i}  \Bigg{\}}.
\end{split}
\end{equation}
If accepted, node $k$ is removed from the active subgraph. 

\paragraph{Moves on order values:} We re-sample the order value of randomly picked node $i$. This operation is done by finding the lowest order valued parent of $i$ along with its highest order valued children, which we respectively denote $l$ and $h$. Next, the candidate order value is sampled according to $\theta_{i} \thicksim \mathcal{U}(\theta_{l}, \theta_{h})$ and accepted with Metropolis-Hasting acceptance ratio: 
\begin{equation}
a_{order} = \min \left\{ 1, \frac{p(Z^\nearrow_{AA}, Z_{IA} , {\boldsymbol\theta'}_{A}^\nearrow | \alpha, \gamma, \phi, O )}{p(Z^\nearrow_{AA}, Z_{IA} , \boldsymbol\theta_A^\nearrow | \alpha, \gamma, \phi, O )}  \right\} \,,
\end{equation}
which proposes a new total order $\boldsymbol\theta$ respecting the partial order imposed by the rest of the DAG. 

\section{Experiments}
\label{sec:learning_experiments}

The ICP distribution \eqref{eq:marginal_joint_proba_infinite_dags} can be used as a prior to learn the structure of any DAG-based model involving hidden units. In particular, one can introduce \textit{a priori} knowledge about the structure by fixing the order values of some observed units. Feedforward neural networks, for instance, can be modelled by imposing $\theta_k = 1$ for all input units and $\theta_k = 0$ for the output units. On the other hand, generative models can be designed by placing all observed units at $\theta_k = 0$, preventing interconnections between them and forcing the above generative units to explain the data. In section \ref{sec:xpgennet}, we use the ICP as a prior to learn the structures of a generative neural network by approximating the full posterior for 9 datasets. In section \ref{sec:xpcnn}, we use the ICP to learn the structure of a convolutional neural network (CNN) in a Bayesian learning framework. 

\subsection{Bayesian nonparametric generative sigmoid network}
\label{sec:xpgennet}

The network used in this section is the Nonlinear Gaussian Belief Network (NLGBN)~\cite{frey1997continuous}, which is basically a generative sigmoid network. In this model, the output of a unit $u_i$ depends on a weighted sum of its parents, where $W_{ki}$ represents the weight of parent unit $u_{k}$, $Z_{ki}$ indicates whether $u_{k}$ is a parent of $u_i$ and $b_i$ is a bias. The weighted sum is then corrupted by a zero mean Gaussian noise of precision $\rho_i$, so that $a_i \thicksim \mathcal{N}(b_i + \sum_{k} Z_{ki} W_{ki} u_{k}, 1/\rho_i)$. The noisy preactivation $a_i$ is then passed through a sigmoid nonlinearity, producing the output value $u_i$. It turns out that the density function of this random output $u_i$ can be represented in closed-form, a property used to form the likelihood function given the data. An ICP prior is placed on the structure represented by $Z$ along with priors $\gamma \thicksim \text{Gamma}(0.5, 0.5)$, $1/\alpha \thicksim \text{Gamma}(0.5, 0.5)$ and $\phi \thicksim \text{Gamma}(0.5, 0.5)$. To complete the prior on parameters, we specify $\rho_k \thicksim \text{Gamma}(0.5, 0.5)$, $b_k \thicksim \mathcal{N}(0,1)$ and $W_{ki} \thicksim \mathcal{N}(0,1)$. 

The inference is done with MCMC where structure operator are given in \ref{sec:structure_learning} and we refer to \cite{adams2010a} for the parameter and activation operators. The Markov chain explores the space of structures by creating and destroying edge and nodes, which means that posterior samples are of varying size and shape, while remaining infinitely layered due to $\theta_k \in [0,1]$. We also simulate the random activations $u_k$ and add them into the chain state. 

This experiment aims at reproducing the generative process of synthetic data sources. In the learning phase, we simulate the posterior distribution conditioned on 2000 training points. Fantasy data from the posterior are generated by first sampling a model from set of posterior network samples and then one point is generated from the selected model. Figure~\ref{fig:results_fantasy} shows 2000 test samples from the true distribution along with the samples generated from the posterior accounting for the model uncertainty.

\newcommand\FC{0.175}

\begin{figure*}[tb]
\centering
\subfigure[groundtruth]{%
    \includegraphics[trim = 10mm 5mm 10mm 5mm, clip,scale=\FC]{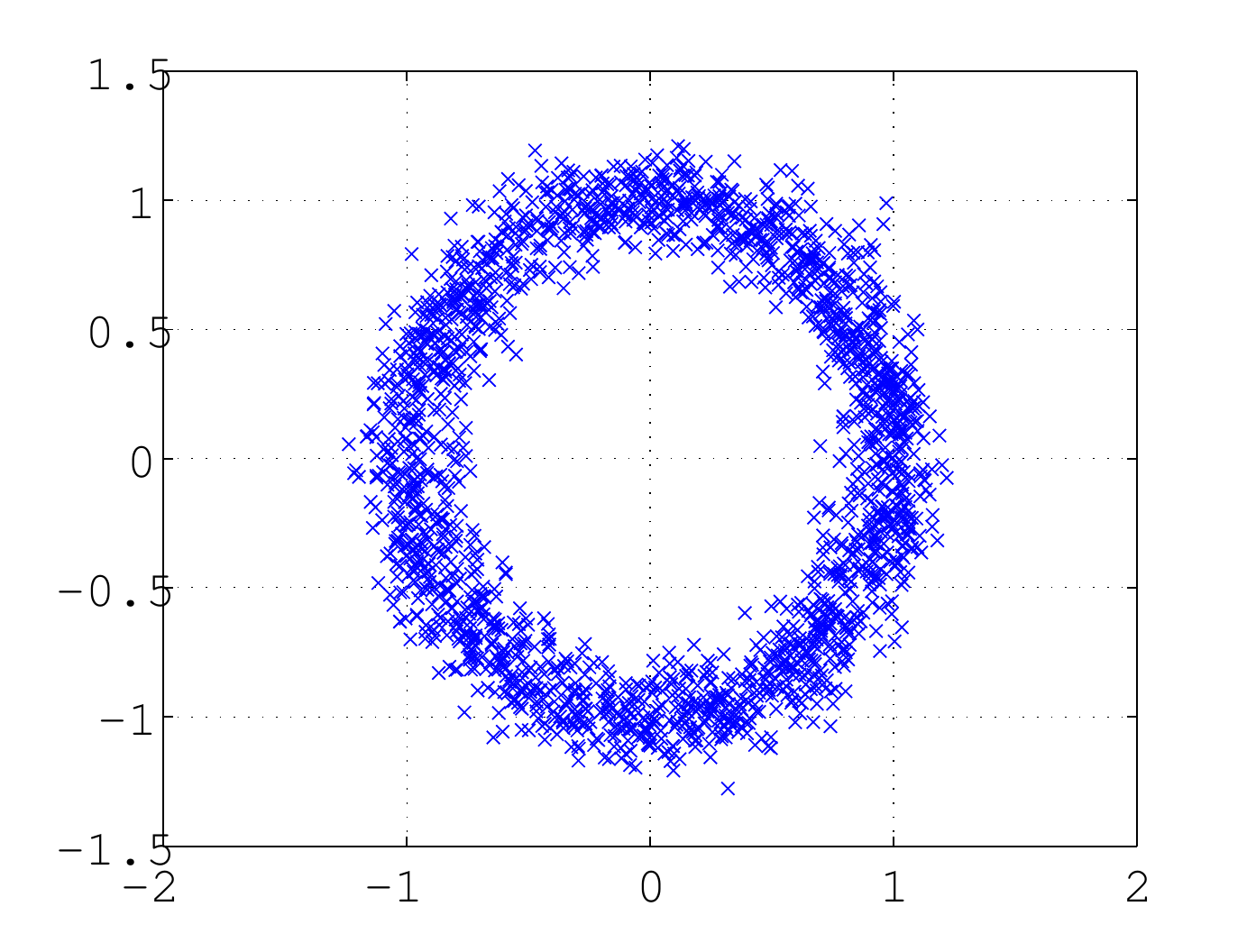}
}
\subfigure[fantasy]{%
    \includegraphics[trim = 10mm 5mm 10mm 5mm, clip,scale=\FC]{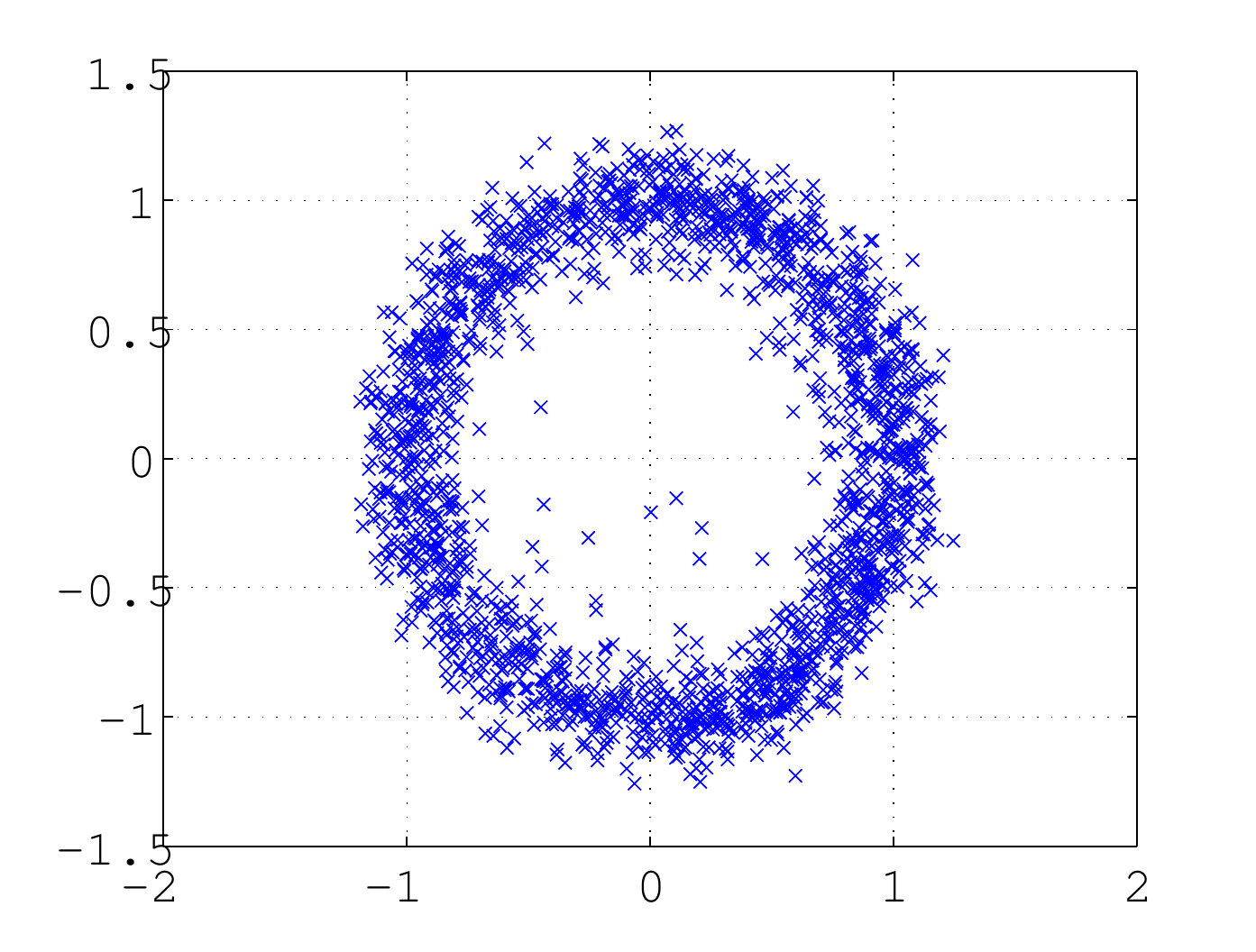}
}
\subfigure[groundtruth]{%
    \includegraphics[trim = 10mm 5mm 10mm 5mm, clip,scale=\FC]{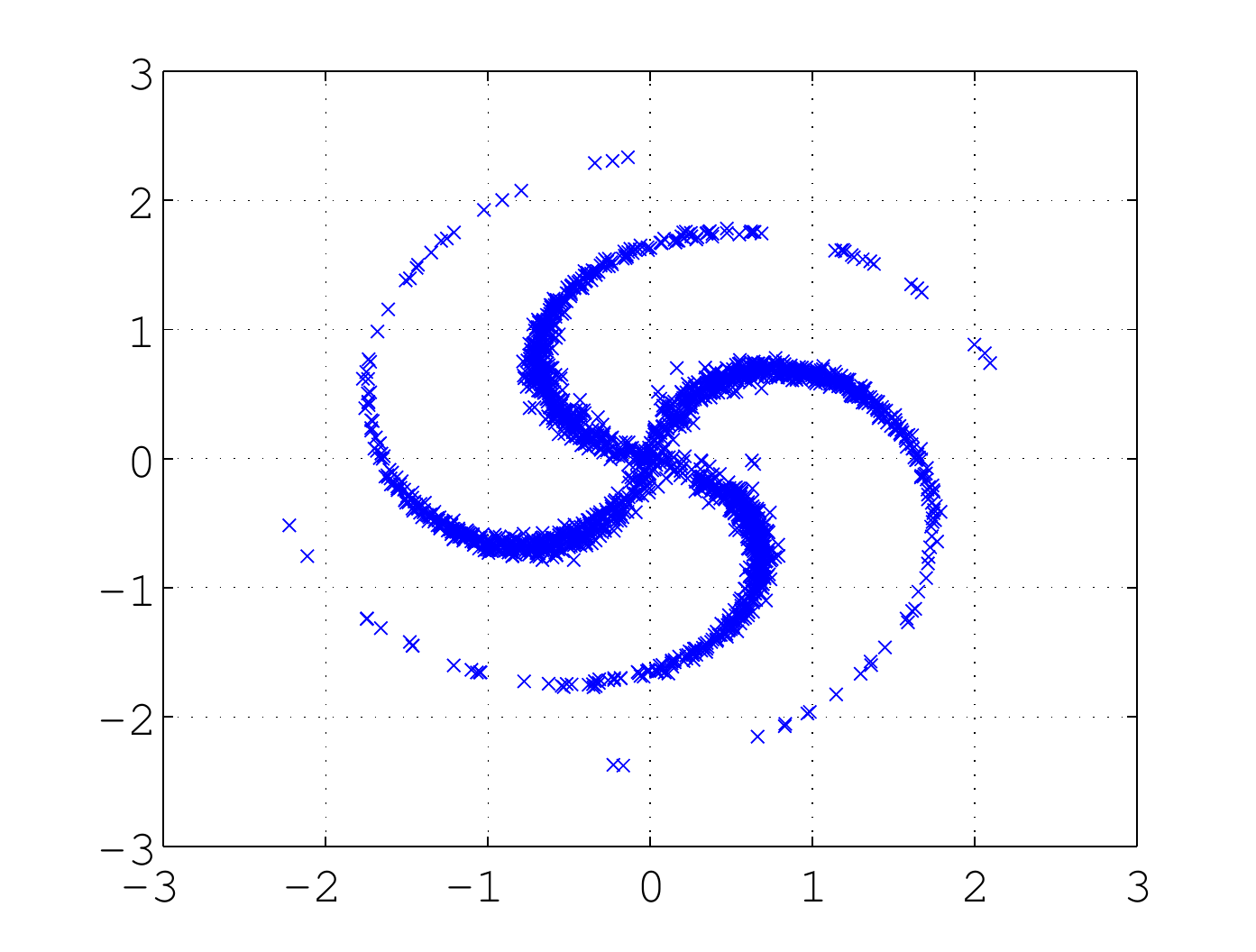}
}
\subfigure[fantasy]{%
    \includegraphics[trim = 10mm 5mm 10mm 5mm, clip,scale=\FC]{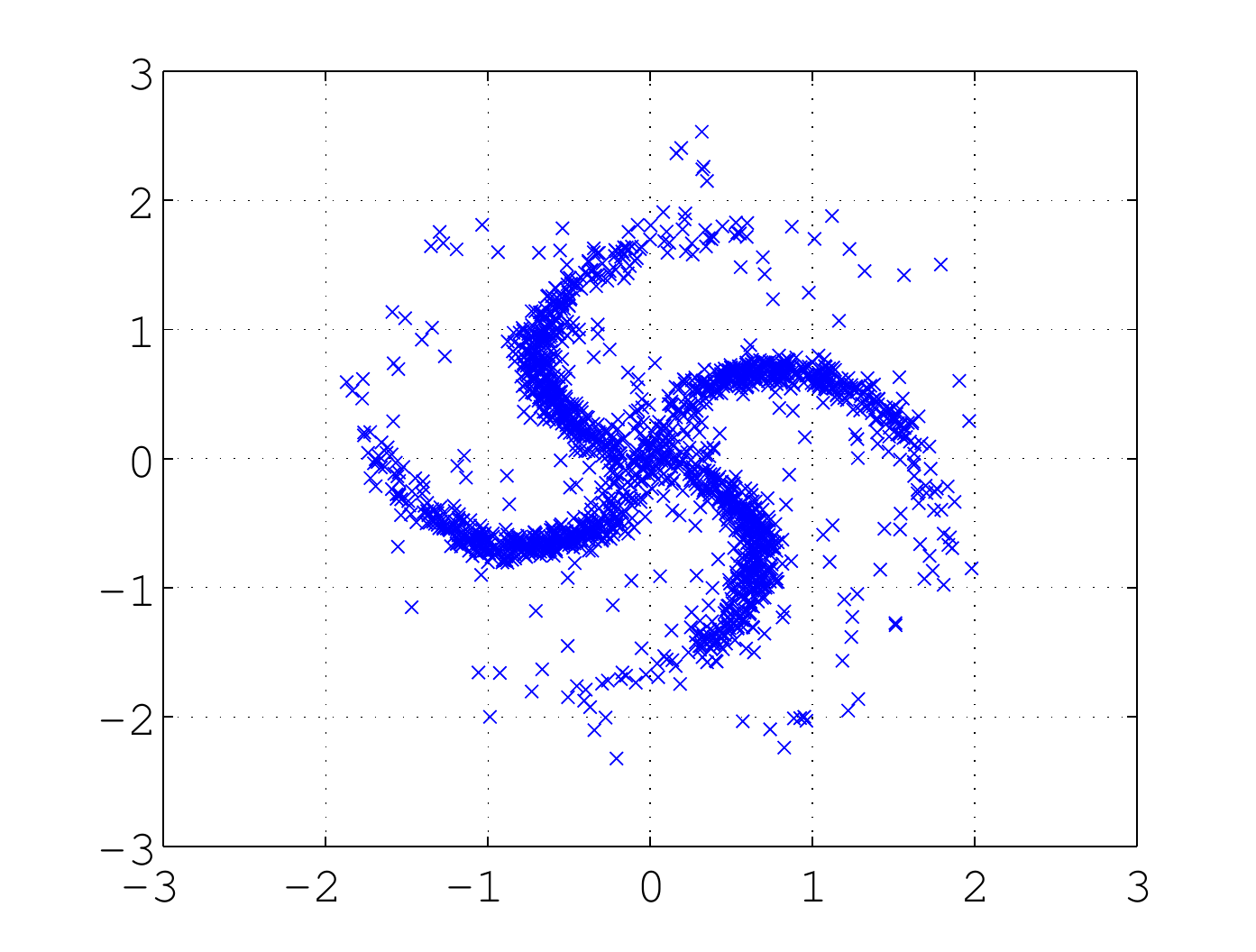}
}
\subfigure[groundtruth]{%
    \includegraphics[trim = 10mm 5mm 10mm 5mm, clip,scale=\FC]{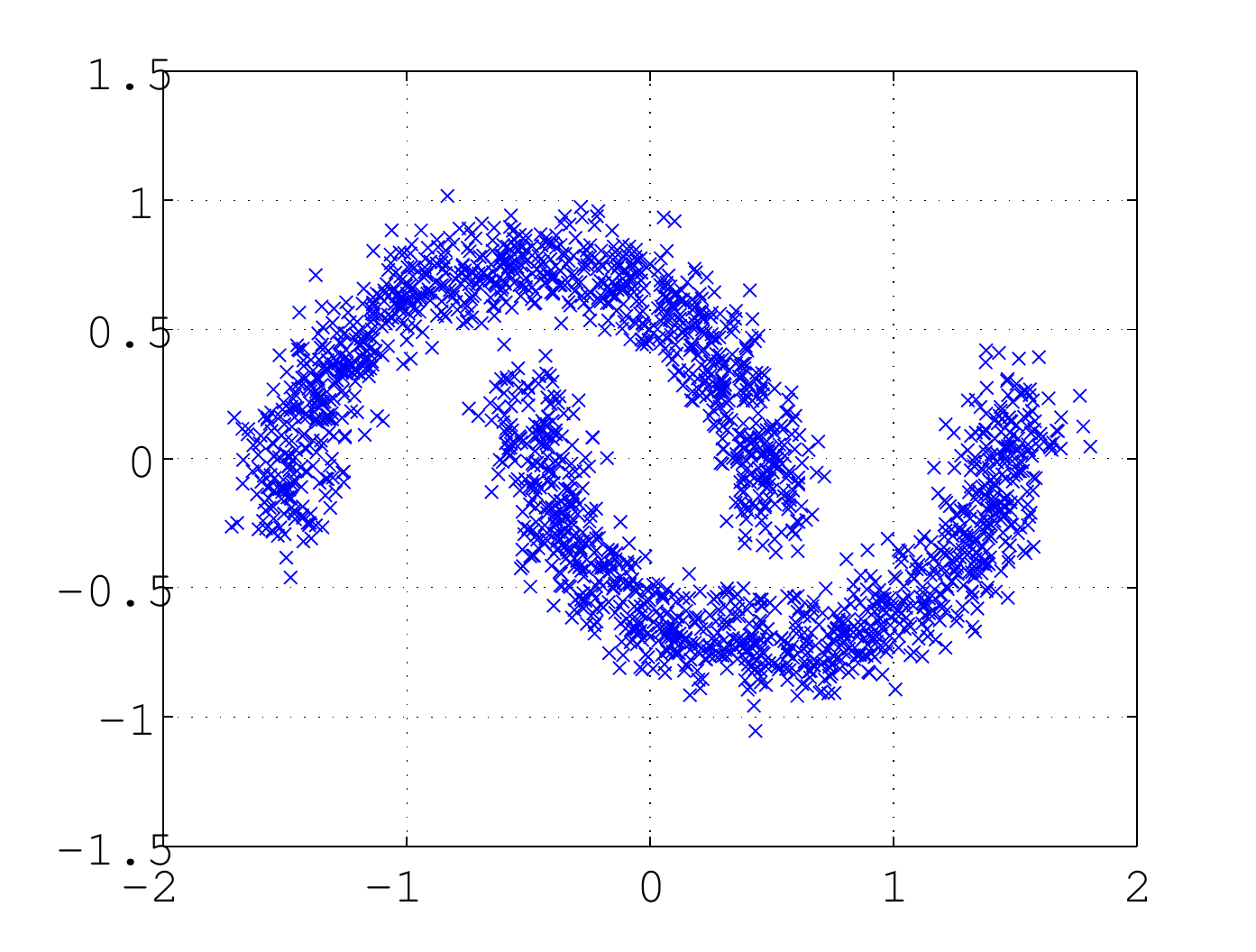}
}
\subfigure[fantasy]{%
    \includegraphics[trim = 10mm 5mm 10mm 5mm, clip,scale=\FC]{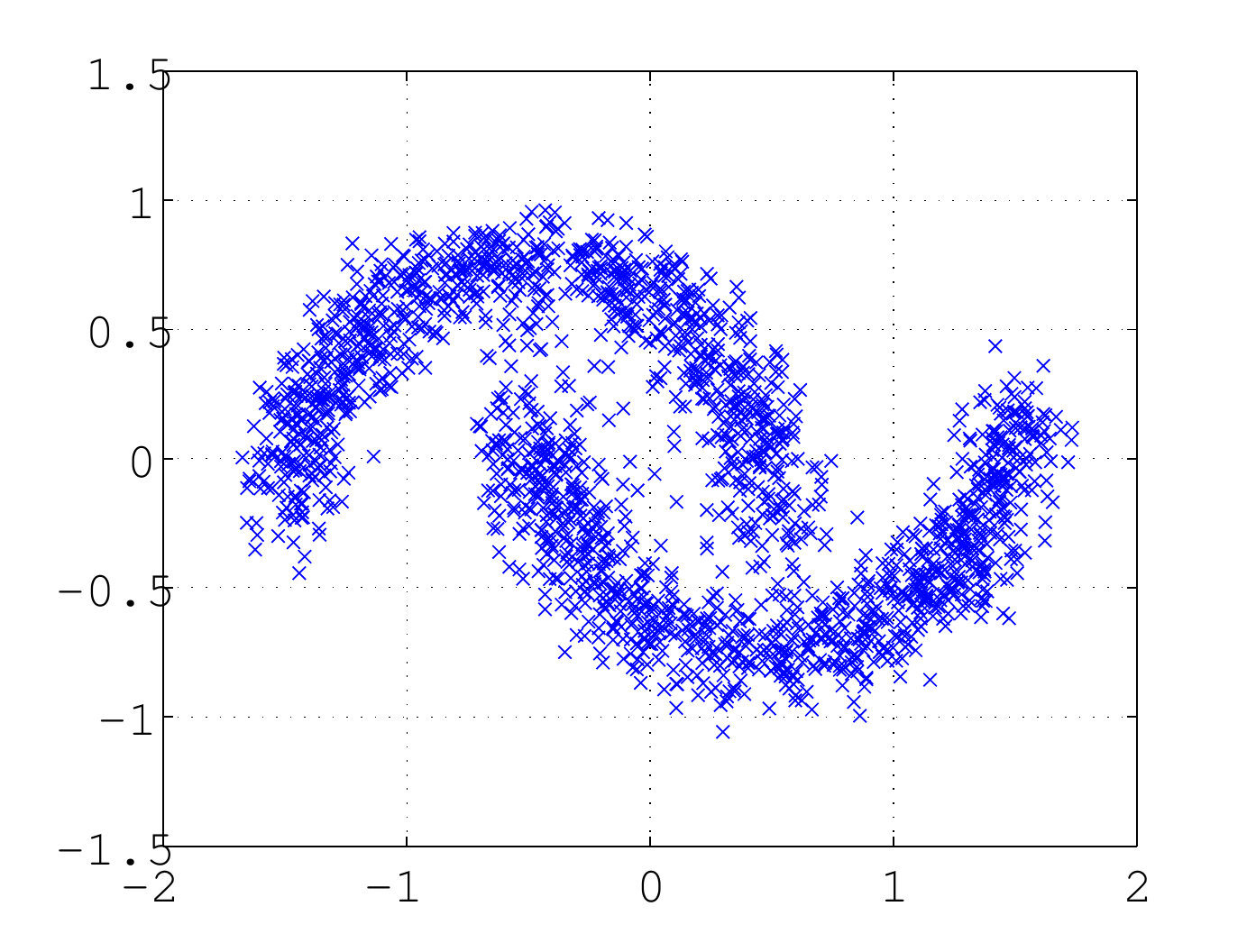}
}
\caption{Resulting fantasy data generated from the posterior on 3 toy datasets.}
\label{fig:results_fantasy}
\end{figure*}

Next, we compare the ICP (with observables at $\theta_k = 0$) against other Bayesian nonparametric approaches: The Cascading Indian Buffet Process~\cite{adams2010a} and the Extended CIBP~\cite{dallaire2014learning}. The inference for these models was done with an MCMC algorithm similar to the one used for the ICP and we used similar priors for the parameters to ensure a fair comparison. The comparison metric used in this experiments is the Hellinger distance~(HD), a function quantifying the similarity between two probability densities. Table~\ref{tab:ehd} shows the 
HD's between fantasy datasets generated and groundtruth.

\begin{table*}[ht!]
{
\caption{Hellinger distance between the fantasy data from posterior models and the test set. Dimensionality of the data is in parenthesis. The baseline shows the distance between the training and test sets, representing the best achievable distance since the two come from the true source.}
\label{tab:ehd}
\tiny
\hfill{}
\begin{tabular}{cccccccccc}
\toprule
Data set& Ring~(2) & Two Moons~(2) & Pinwheel~(2) & Geyser~(2) & Iris~(4) & Yeast~(8) & Abalone~(9) &  Cloud~(10) & Wine~(12) \\
\midrule
ICP & \textbf{0.0402} & \textbf{0.0342} & \textbf{0.0547} & \textbf{0.0734} & 0.2666 & \textbf{0.3817} & \textbf{0.1379} & \textbf{0.1495} & \textbf{0.3629} \\
CIBP & 0.0493 & 0.0469 & 0.0692 & 0.1246 & 0.2667 & 0.4056 & 0.1502 & 0.1713 & 0.4079 \\
ECIBP & 0.0419 & 0.0450 & 0.0685 & 0.1171 & \textbf{0.2632} & 0.3840 & 0.1470 & 0.1501 & 0.3855 \\
\midrule
Baseline & 0.0312 & 0.0138 & 0.0436 & 0.0234 & 0.1930 & 0.3059 & 0.1079 & 0.1299 & 0.3387 \\
\bottomrule
\end{tabular}
}
\hfill{}
\end{table*}

\subsection{Bayesian nonparametric convolutional neural networks}
\label{sec:xpcnn}

So far, we introduced the ICP as a prior on the space of directed acyclic graphs. In this section we will use this formalism in order to construct a prior on the space of convolutional neural architectures. The fundamental building blocks of (2D) convolutional networks are tensors $T$ whose entries encode the presence of local features in the input image. A convolutional neural network can be described as a sequence of convolution operators acting on these tensors followed by entry-wise nonlinearity $f$.

In our nonparametric model, a convolutional network is constructed from a directed acyclic graph. Each node of the graph represents a tensor $T^{(i)}$. The entries of this tensor are given by the following:
\begin{equation}
    T^{(i)} = \text{ReLu}\! \left(\sum_{k \in \text{Parents}(i)} W^{(ki)} \star T^{(k)} \right)~.
\end{equation} where $W^{(ki)}$ is a tensor of convolutional weights and $\star$ is the discrete convolution operator.  In most hand-crafted architectures, the spatial dimensions of the tensor are course-grained as the depth increases while the number of channels (each representing a local feature of the input) increases. In the ICP, the depth of a node $i$ is represented by its popularity $\theta_i$. In order to encode the change of shape in the nonparametric prior, we set the number of channels to be a function of $\theta$:
\begin{equation}
    N_c(\theta) =  2^{\lfloor N_{\text{bins}} (1 - \theta))  \rfloor} + N_0  ~, 
\end{equation}
where $N_\text{bins}$ is the number of different possible tensor shapes and $N_0$ is the number of channels of the lowest layers. Similarly, the number of pixels is given by:
\begin{equation}
    N_p(\theta) =  2^{-\lfloor N_{\text{bins}} (1 - \theta))  \rfloor} M  ~, 
\end{equation}
where $M$ is the number of pixels in the original image. The shape of the weight tensors $W^{(ki)}$ is determined by the shape of parent and child tensor.

In a classification problem, the nonparametric convolutional network is connected to the data through two observed nodes. The input node $X$ stores the input images and we set $\theta_X = 1$. On the other hand, for the output node we set $\theta_Y=0$, and have it receive input through fully connected layers:
\begin{equation}
    Y = \text{SoftMax}\! \left(\sum_{k \in \text{Parents}(Y)} \left( \sum_{a,b,c} V^{(k)}_{a} T^{(k)}_{abc} \right) \right)~,
\end{equation}
where $V^{(k)}$ is a tensor of weights. During sampling of a connection $Z_{ki}$ in the directed acyclic graph, the log acceptance ratio needs to be summed to the log marginal-likelihood ratio:
\begin{equation}
    \log{a_{\text{lk}}} =  \log{p(y\mid DAG, X)} - \log{p(y\mid DAG_{\text{proposal}}, X)}~.
\end{equation}
In this paper, we use a point estimate of the log model evidence:
\begin{equation}
    p(y \mid DAG, x) \approx p(y\mid DAG, \{\hat{W}^{(ki)} \}, x)~,
\end{equation}
where $\hat{W}^{(jk)}$ are the parameters of the network optimized using Adam (alpha=0.1, beta1=0.9, beta2=0.999, eps=1e-08, eta=1.0) \cite{kingma2014adam}.

We performed architecture sampling on the MNIST dataset. For computational reasons, we restricted the dataset to the first three classes (1,2 and 3). We sampled the DAGs using the MCMC sampler introduced in section~\ref{sec:structure_learning} with prior parameters $\alpha = 1$, $\gamma = 20$ and $\phi = 5$. For each sampled DAG, we trained the induced convolutional architecture until convergence (540 iterations, batch size equal to 100). The number of bins in the partition of the range of the popularity was five and the number of channels of the first convolutional layer was four. We ran 15 independent Markov chains in order to sample from multiple modes. Each chain consisted of 300 accepted samples. After sampling, all chains were merged, resulting in a total of 4500 sampled architectures.

\begin{figure} 
\centering
\includegraphics[width=0.50\textwidth]{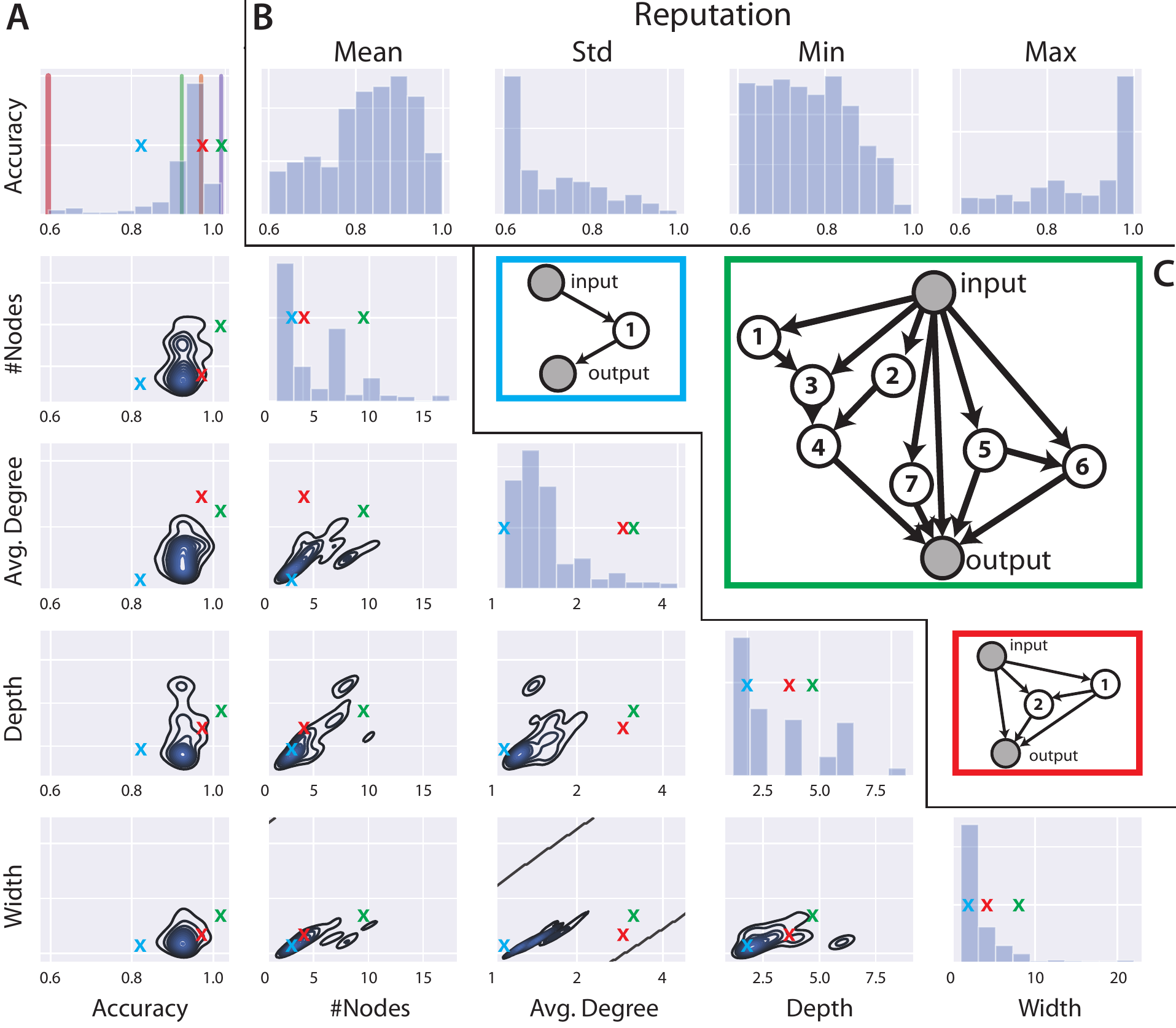}
\caption{Statistics of the sampled convolutional architectures. A) Histograms and bivariate density plots of test set accuracy, number of nodes, average degree width and depth. The three colored crosses denote the statistics of the three visualized networks. B) Histogram of the mean, standard deviation, min and max of the popularity values.}
\label{fig:results_convolutional}
\end{figure}

Figure~\ref{fig:results_convolutional}A shows accuracy and descriptive statistics of the sampled convolutional architectures. In all these statistics, we only considered nodes that receive input from the input node (directly or indirectly) as the remaining nodes do not contribute to the forward pass of the network. The network \emph{width} is quantified as the total number of directed paths between input and output nodes, while \emph{depth} is quantified as the maximal directed path length. The sampler reaches a wide range of different architectures, whose number of layers range from three to fifteen, and whose average degree range from one to four. Some examples of architectures are shown in Figure~\ref{fig:results_convolutional}C. Interestingly, the correlation between the number of nodes, degree, width and depth and accuracy is very low. Most likely, this is due to the simple nature of the MNIST task. The ensemble accuracy (0.95), obtained by averaging the label probabilities over all samples, is higher that the average accuracy (0.91), but lower that the maximum accuracy (0.99). Figure~\ref{fig:results_convolutional}B shows the histograms of mean, standard deviation, minimum and maximum of the popularity values in the networks.

\section{Conclusion and Future Work}
\label{sec:discussion}

This paper introduces the Indian Chefs Process (ICP), a Bayesian nonparametric distribution on the joint space of infinite directed acyclic graphs and orders. The model allows for a novel way of learning the structure of deep learning models. As a proof-of-concept, we have demonstrated how the ICP can be used to learn the architecture of convolutional deep networks trained on the MNIST data set. However. for more realistic applications, several efficiency improvements are required. Firstly, the inference procedure over the model parameters could be performed using stochastic Hamiltonian MCMC~\cite{ma2015complete}. This removes the need to fully train the network for every sampled DAG. Another possible improvement is to add deep learning specific sampling moves. For example, it is possible to include a increase-depth move that replaces a connection with a path comprised by two connections and a latent node. Future applications of the ICP may extend beyond deep learning architectures. For example, the ICP may serve as a basis for nonparametric causal inference, where a DAG structure is learned when the exact number of relevant variables is not known a priori, or when certain relevant input variables are not observed~\cite{Mohan2013}.

\bibliographystyle{abbrv}
\bibliography{bibliography.bib}

\end{document}